\documentclass[pmlr]{jmlr}

\usepackage{longtable}
\usepackage{subcaption}
\usepackage{hyperref}
\usepackage[croatian]{babel}

\usepackage{booktabs}
\usepackage[load-configurations=version-1]{siunitx} 

\theorembodyfont{\upshape}
\theoremheaderfont{\scshape}
\theorempostheader{:}
\theoremsep{\newline}

\jmlrvolume{1}
\jmlryear{2019}
\jmlrworkshop{NeurIPS2019 Disentanglement Challenge}

\title[Disentanglement-PyTorch]{Variational Learning with Disentanglement-PyTorch}

  \author{\Name{Amir H. Abdi} \Email{amirabdi@ece.ubc.ca}\\
   \Name{Purang Abolmaesumi} \Email{purang@ece.ubc.ca}\\
   \Name{Sidney Fels} \Email{ssfels@ece.ubc.ca}\\
   \\
   \addr Electrical and Computer Engineering Department, University of British Columbia, Canada}

\begin{document}

\maketitle

\begin{abstract}
Unsupervised learning of disentangled representations is an open problem in machine learning.
The Disentanglement-PyTorch library is developed to facilitate research, implementation, and testing of new variational algorithms.
In this modular library, neural architectures, dimensionality of the latent space, and the training algorithms are fully decoupled, allowing  for independent and consistent experiments across variational methods. 
The library handles the training scheduling, logging, and visualizations of reconstructions and latent space traversals. 
It also evaluates the encodings
based on various disentanglement metrics. 
The library, so far, includes implementations of the following unsupervised algorithms
VAE, $\beta$-VAE, Factor-VAE, DIP-I-VAE, DIP-II-VAE, Info-VAE, and $\beta$-TCVAE, 
as well as conditional approaches such as CVAE and IFCVAE.
The library is compatible with the Disentanglement Challenge of NeurIPS 2019, hosted on AICrowd, and achieved the 3rd rank in both the first and second stages of the challenge. 


\end{abstract}

\section{Introduction}
\label{sec:intro}

There are two overlapping avenues in representation learning. 
One  focuses on learning task-specific transformations often optimized towards specific domains and applications.
The other approach learns the intrinsic factors of variation, in a disentangled and task-invariant fashion.
The unsupervised disentanglement of latent factors, 
where changes in a single factor of variation shifts the latent encoding in a single direction,
is an open problem of representation learning~\citep{Bengio2013,Lake2016}. 
Disentangled representations are valuable in few-shot learning, reinforcement learning, transfer learning, as well as semi-supervised learning~\citep{Bengio2013,PetersJanzingSchoelkopf17,GoogleAssumptions}.


In this work, we developed a library based on the functionalities of the PyTorch framework, which facilitates research, implementation, and testing of new variational algorithms focusing on representation learning and disentanglement.
The library branches from the Disentanglement Challenge of NeurIPS 2019, hosted on AICrowd (\href{https://www.aicrowd.com}{aicrowd.com}), and was used to compete in the first and second stages of the challenge where it was highly ranked.
The Disentanglement-PyTorch library is released under the GNU  General Public License at \url{https://github.com/amir-abdi/disentanglement-pytorch}.

\section{Library Features}

\subsection{Supported Algorithms and Objective Functions}
\label{sec:algs}

\paragraph{Unsupervised Objectives}
Currently, the library includes implementations of the following unsupervised variational algorithms:
VAE~\citep{VAE}, $\beta$-VAE~\citep{BetaVAE}, $\beta$-TCVAE~\citep{betatcvae}, Factor-VAE~\citep{FactorVAE}, Info-VAE~\citep{InfoVAE}, DIP-I-VAE,  and DIP-II-VAE~\citep{DIPVAE}.
Algorithms are implemented as plug-ins to the variational Bayesian formulation, and are specified by the \texttt{loss\_terms} flag.
As a result, if the loss terms of two learning algorithms (\emph{e.g.}, A and B) were found to be compatible, they can both be included in the objective function with the flag set as [\texttt{--loss\_terms A B}].
This enables researchers to mix and match loss terms which 
optimize towards correlated goals.

\paragraph{Conditional and Attribute-variant Objectives}
The library supports conditional approaches such as CVAE~\citep{CVAE}, where extra known attributes (\emph{i.e}, labels) are
included in the encoding and decoding processes.
It also supports IFCVAE,  inspired by the IFcVAE-GAN~\citep{IFCVAE}, which 
enforces certain latent factors to encode known attributes using a set of positive (auxiliary) and negative (adversarial) discriminators in a supervised fashion.
Thanks to the modular implementation of the library, 
any of the above-mentioned unsupervised loss terms 
can be used with  conditional and information factoriation 
approaches to encourage disentanglement across attribute-invariant latents.

\subsection{Neural Architectures}

Neural architectures and the dimensionality of the data and the latent spaces are configurable and decoupled from the training algorithm.
Consequently, new architectures for the encoder and decoder networks, 
such as the auto-regressive models,
and support for  other data domains, can be independently investigated.

\subsection{Evaluation of Disentanglement}

We rely on Google's implementation of the disentanglement metrics to  evaluate the quality of the learned representations~\citep{GoogleAssumptions}.
Thanks to the 
\href{https://github.com/google-research/disentanglement_lib}{disentanglement-lib}\footnote{\url{https://github.com/google-research/disentanglement_lib}}
library, the following metrics are currently supported:
BetaVAE~\citep{Higgins2017betaVAELB}, 
FactorVAE~\citep{FactorVAE},
Mutual Information Gap (MIG)~\citep{betatcvae},
Interventional Robustness Score (IRS)~\citep{IRS},
Disentanglement Completeness and Informativeness (DCP)~\citep{DCIMetric}, and
Separated Attribute Predictability (SAP)~\citep{DIPVAE}.

\subsection{Miscellaneous Features}

\paragraph{Controlled Capacity Increase} 
It is shown that gradually relaxing the information bottleneck during training improves the disentanglement without penalizing the reconstruction accuracy.
Following the formulation of \cite{Capacity},
the capacity, defined as the distance between the prior and the latent posterior distributions and denoted with the variable $C$, is gradually increased during training.


\paragraph{Reconstruction Weight Scheduler} 

To avoid convergence points with  high reconstruction loss, 
training can be started with more emphasis on the reconstruction and gradually relaxing for the disentanglement term to become more relative.

\paragraph{Dynamic Learning Rate Scheduling}
All forms of learning rate  schedulers are supported.
Researchers are 
encouraged  to leverage
the dynamic LR scheduling to gradually decrease the rate when the 
average objective function over the  epoch stops its decremental trend.

\paragraph{Logging and Visualization}
The library leverages the Weights \& Biases\footnote{https://www.wandb.com} tool to log the training process and the visualizations.
It visualizes the condition's traversals, latent factor traversals, 
and Output reconstructions as \href{https://app.wandb.ai/amirabdi/disentanglement-pytorch/runs/8k9dsisu?workspace=user-amirabdi}{static images} and \href{https://github.com/amir-abdi/disentanglement-pytorch/blob/master/sample_results/mpi3d_realistic_VAE/gif_rand0.gif?raw=true}{animated GIFs}.

\section{Experiments and Results}

The $\beta$-TCVAE algorithm  achieved the best disentanglement results on the \texttt{mpi3d\_real} dataset in the second stage of the disentanglement challenge.
Given the limited 8-hour training time for the challenge, 
the model was pre-trained on the \texttt{mpi3d\_toy} dataset~\citep{mpi3d}.
The model was trained with the Adam optimizer for 90k iterations on batches of size 64.
The  $\beta$ value of the $\beta$-TCVAE objective function was set to 2.
The learning rate was initialized at 0.001 and reduced on the plateau of the objective function with a factor of 0.95.
The capacity parameter, $C$, was gradually increased from 0 to 25.
The dimensionality of the $z$ space was generously set to 20.

The encoder consisted of 5 convolutional layers with strides of 2, kernel sizes of $3\times3$, and number of kernels gradually increasing from 32 to 256. 
The encoder ended with a dense linear layer which estimated the posterior latent distribution as a parametric Gaussian.
The decoder network consisted of one convolutional followed with 6 deconvolutional (transposed convolutional) layers, with kernel sizes of 4, strides of 2, and the number of kernels gradually decreasing from 256 down to the number of channels of the image space. 
ReLU activations were used except for the last layers of the encoder and decoder networks.

Model's performance on the unseen objects of \texttt{mpi3d\_realistic} and \texttt{mpi3d\_real} datasets are presented in Table~\ref{tab:results}.
The configurations of the two experiments (stage 1 and 2) were the same and the model consistently performed better on the \texttt{mpi3d\_realistic} and \texttt{mpi3d\_real} datasets.
This was unexpected as the model was only pretrained on the \texttt{mpi3d\_toy} dataset.
Qualitative results of disentanglement training on the available samples of the \texttt{mpi3d\_realistic} dataset
are visualized in Appendix~\ref{apd:traverse}.



\begin{table}[hbtp]
\floatconts
  {tab:results}
  {\caption{Results of the best configurations of $\beta$-TCVAE on DCI, FactorVAE, SAP, MIG, and IRS metrics.}}
  {\begin{tabular}{lcccccc}
  \toprule
  \bfseries Method & Dataset & \bfseries DCI & \bfseries FactorVAE  & \bfseries  SAP & \bfseries MIG & \bfseries IRS \\
  \midrule
  $\beta$-TCVAE & \texttt{mpi3d\_realistic} &  0.3989 &	0.3614	& 0.1443 & 	0.2067 & 0.6315 
\\
$\beta$-TCVAE & \texttt{mpi3d\_real} &  0.4044 &	0.5226	& 0.1592 & 	0.2367 &  0.6423
\\
  \bottomrule
  \end{tabular}}
\end{table}

\bibliography{jmlr-sample}

\pagebreak
\appendix

\section{Latent Factor Traversal}\label{apd:traverse}

\begin{figure}[!ht]
\floatconts
  {fig:beta-tcvae}
  {\caption{Latent factor traversal of the trained  \textbf{$\beta$-TCVAE} model on a random sample of the \texttt{mpi3d\_realistic} dataset.
  As demonstrated, the disentanglement is not complete and some features are encoded in the same latent factor.
  A latent space of size 20 was used, however, changes in the other 13 latent factors had no effect on the reconstruction; thus, these feature-invariant factors were not included for brevity.}}
  {\includegraphics[width=1.0\linewidth]{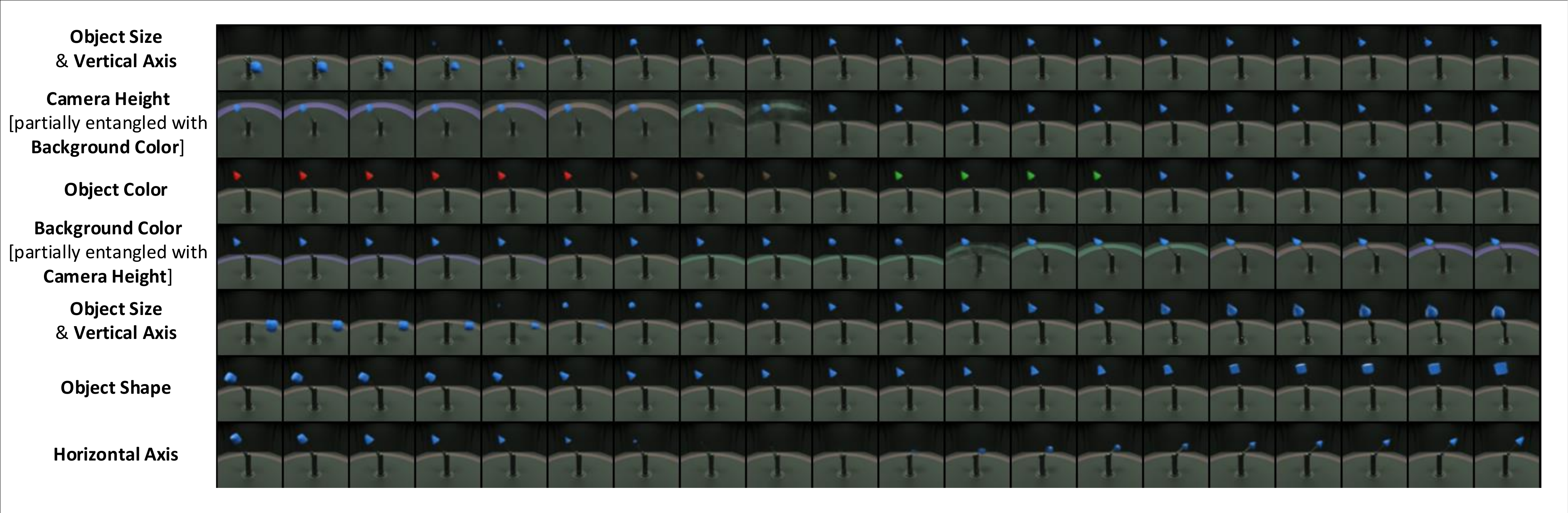}}
\end{figure}

\end{document}